\documentclass[transmag]{IEEEtran}
\usepackage{latexsym}
\usepackage{graphicx}
\usepackage{amsfonts,amssymb,amsmath}
\usepackage{hyperref}
\usepackage{algorithm}
\usepackage{algpseudocode}
\usepackage{amsmath}
\usepackage{amssymb}
\usepackage{amsmath,epsfig,bm,subfigure}
\usepackage{multirow}
\usepackage{tabu}
\usepackage{makecell}
\usepackage{cite}
\usepackage{threeparttable}
\usepackage{url}
\usepackage{booktabs,bigstrut}
\usepackage{flushend}
\usepackage{color}

\def\BibTeX{{\rm B\kern-.05em{\sc i\kern-.025em b}\kern-.08em T\kern-.1667em\lower.7ex\hbox{E}\kern-.125emX}}

\ifCLASSINFOpdf

\else

\fi

\begin{document}

\title{Learned Image Compression with Separate Hyperprior Decoders}

\author{Zhao~Zan,
        Chao~Liu,
        Heming~Sun,
        Xiaoyang~Zeng,
        and~Yibo~Fan

\thanks{
This work was supported in part by the National Natural Science Foundation of China under Grant 62031009, in part by the Shanghai Science and Technology Committee (STCSM) under Grant 19511104300, in part by Alibaba Innovative Research (AIR) Program, in part by the Innovation Program of Shanghai Municipal Education Commission, in part by the Fudan University-CIOMP Joint Fund (FC2019-001), in part by the Fudan-ZTE Joint Lab, in part by JST, PRESTO Grant Number JPMJPR19M5, Japan. \textit{(Corresponding author: Heming Sun and Yibo Fan.)}}

\thanks{H. Sun is with the Waseda Research Institute for Science and Engineering, Tokyo 169-8555, Japan and JST, PRESTO, 4-1-8 Honcho, Kawaguchi, Saitama, 332-0012, Japan (e-mail: hemingsun@aoni.waseda.jp).}

\thanks{Y. Fan is with the State Key Laboratory of ASIC and System, Fudan University, Shanghai 200433, China (e-mail: fanyibo@fudan.edu.cn).}
}

\IEEEtitleabstractindextext{\begin{abstract}
Learned image compression techniques have achieved considerable development in recent years. In this paper, we find that the performance bottleneck lies in the use of a single hyperprior decoder, in which case the ternary Gaussian model collapses to a binary one. To solve this, we propose to use three hyperprior decoders to separate the decoding process of the mixed parameters in discrete Gaussian mixture likelihoods, achieving more accurate parameters estimation.
Experimental results demonstrate the proposed method optimized by MS-SSIM achieves on average 3.36\% BD-rate reduction compared with state-of-the-art approach. The contribution of the proposed method to the coding time and FLOPs is negligible.
\end{abstract}


}

\maketitle

\section{INTRODUCTION}

\IEEEPARstart{I}{mage} compression is an essential technology in digital age. Traditional codecs\cite{jpeg,webp,avif,bpg,hevc,vvc}, such as JPEG\cite{jpeg}, BPG\cite{bpg} and VVC\cite{vvc} have achieved significant coding efficiency. However, as the design complexity and coding complexity continue to increase, it becomes increasingly difficult to further optimize them.
In addition, modules in traditional codecs designed with the optimization goal of minimizing mean square error (MSE) also make it difficult to optimize for general quality evaluation metrics.

With the resurgence of artificial neural network techniques, learned image codecs \cite{balle2017end, balle2018variational, lee2018context, minnen2018joint, cheng2020learned, guo2021soft, guo2021accelerate, toderici2017full, johnston2018improved, agustsson2019generative, mentzer2020high, ma2019iwave, ma2020end} have attracted wide interest in recent years.
By jointly optimizing distortion and rate through Lagrangian multiplication, the work \cite{balle2017end} have developed a framework for end-to-end training of image compression model and achieved impressive performance.
Based on this model, researchers have carried out extensive efforts \cite{balle2018variational, lee2018context, minnen2018joint, cheng2020learned} to reduce redundancy in the latent variables.
Balle \textit{et al.} \cite{balle2018variational} proposed a hyperprior network based on variational autoencoder that consumes a small number of extra bits to encode the structural information of the latent representation.
Lee \textit{et al.} \cite{lee2018context} and Minnen \textit{et al.} \cite{minnen2018joint} proposed the use of context models to further reduce the spatial correlation in the latent space.
Cheng \textit{et al.} \cite{cheng2020learned} proposed using a Gaussian mixture likelihood to parameterize the distributions of latent variables, providing more flexibility to fit arbitrary distributions.
Guo \textit{et al.} \cite{guo2021soft} achieved train-test consistency and reserved latent expressiveness via a novel soft-then-hard quantization method.
Guo \textit{et al.} \cite{guo2021accelerate} utilized a channel-adaptive codebook to accelerate arithmetic coding of learned image compression while maintaining the rate-distortion performance.
In addition to the variational autoencoder-based methods, a substantial body of work based on other learned structures have also achieved impactful results. Recurrent neural networks-based methods \cite{toderici2017full, johnston2018improved} have good scalability in coding and can recursively compress the residual information.
Generative adversarial networks-based approaches \cite{agustsson2019generative, mentzer2020high} are able to achieve excellent subjective quality at extremely low rates.
Flow-based model \cite{ma2019iwave, ma2020end} allows a single model to achieve both lossy and lossless compression of images through a wavelet-like transform and optional quantization, which is potential to surpass variational autoencoder based methods.

\begin{figure}[tbp]
    \centering     
    \subfigure[Mixed hyperprior decoder]{\label{fig1a}\centering  \includegraphics[scale=0.45]{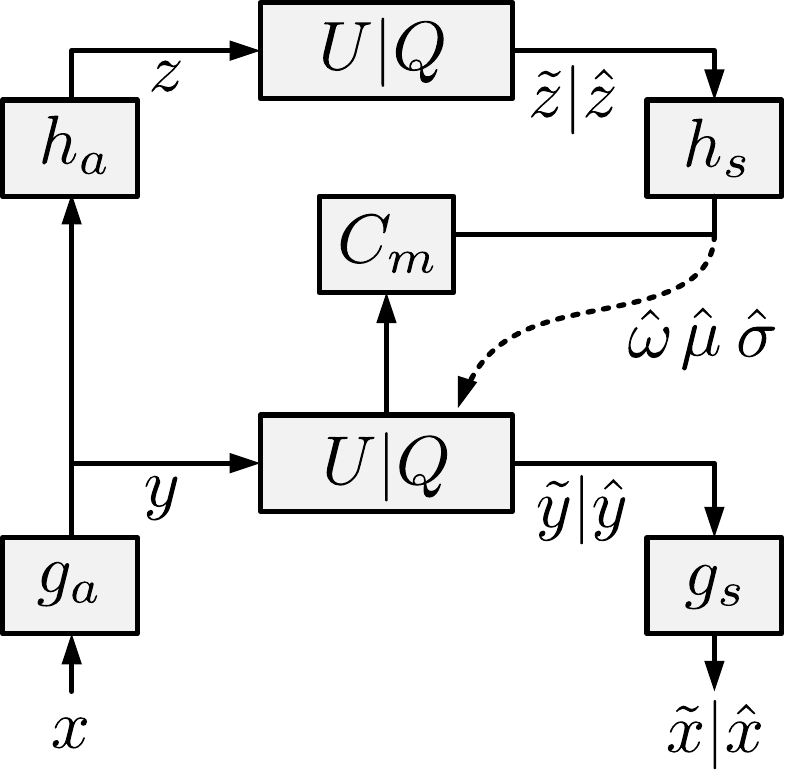}}
    \subfigure[Separate hyperprior decoders]{\label{fig1b}\centering  \includegraphics[scale=0.45]{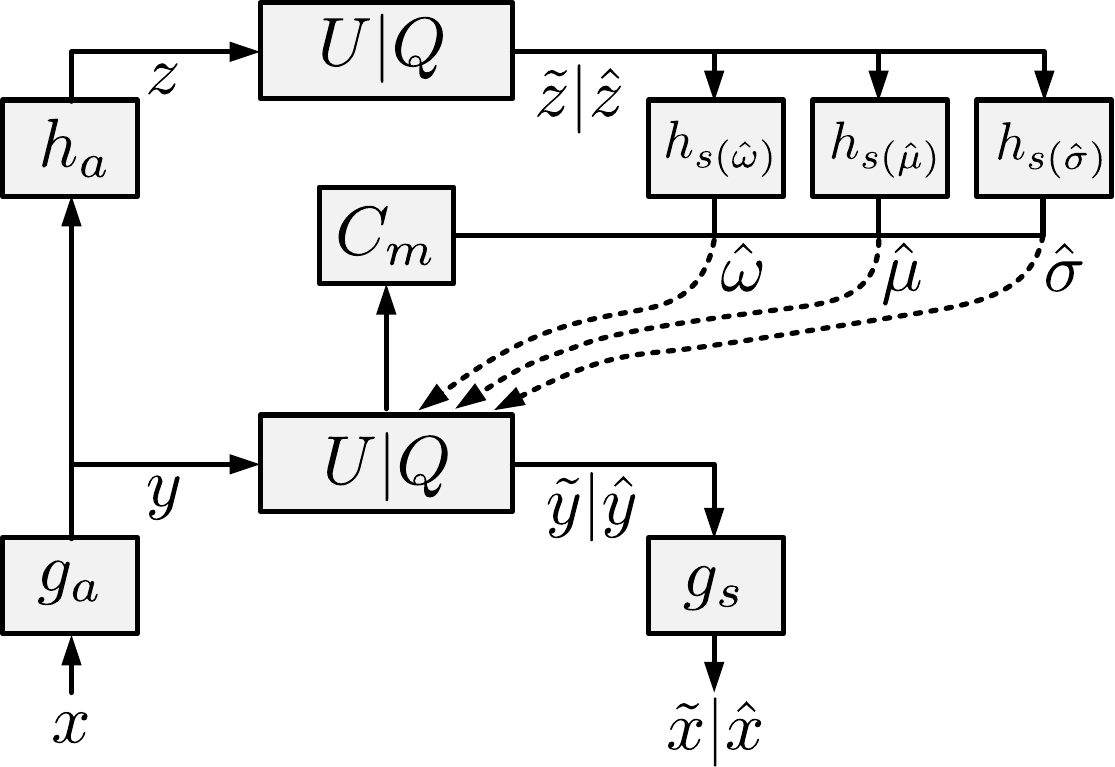}}
    \caption{Operational diagrams of different hyperprior decoder structures for learned compression framework.\label{fig1}}
\end{figure}

In this paper, we focus on the variational autoencoder-based image compression framework. Instead of directly minimizing the redundancy in the latent space, we employ a direct and effective structure to obtain higher compression performance.
The work we study in this paper is based on the work of Cheng \textit{et al.} \cite{cheng2020learned}, which uses a Gaussian mixture model (GMM) prior and achieved state-of-the-art performance.
In Cheng \textit{et al.}\cite{cheng2020learned}, as shown in Fig. \ref{fig1a}, the decoding process of GMM parameters uses only a single hyperprior decoder, which leads to the inability to fully exploit the GMM's ability to fit the data and becomes a bottleneck that constrains the compression performance.
When we use a single hyperprior decoder, the decoding process of the three parameters, mean, variance, and weight in Gaussian model, needs to share the same hyper decoder.
Decoding three different physically significant parameters simultaneously may be somewhat difficult for a single decoder, and the weights of the final trained decoder are a compromise of the three.
We perform an intuitive demonstration to show that this leads to degradation from a ternary Gaussian model to a binary one.
In order to avoid this problem, we propose separate hyperprior decoders as shown in Fig. \ref{fig1b}, to decouple the parameters of different physical significant in GMM and accordingly design different decoding networks to train and decode the parameters of the likelihood distribution more efficiently.

\section{Proposed Method}
\subsection{Formulation of Learned Image Compression}
The image compression process based on the variational autoencoder \cite{balle2017end} can be formulated by
\begin{align}\label{eq1}
  \bm{y} & =g_a(\bm{x};\bm{\phi}) \notag \\
  \bm{\hat{y}} &=Q(\bm{y}) \\
  \bm{\hat{x}} & =g_s(\bm{\hat{y}};\bm{\theta}) \notag
\end{align}
where $\bm{x},\bm{\hat{x}},\bm{y},\bm{\hat{y}}$ denote input images, reconstructed images, the latent variables and the quantized latent variables, respectively. Notation $Q$ denotes real round-based quantization in inference stage. Notation $g_a$ and $g_s$ denote the encoder and decoder, respectively, and $\phi$ and $\theta$ correspond to their parameters. In the training process, considering that non-differentiable quantization will result in the inability to back-propagate the gradient, the work uses a uniform noise to replace the quantization here.
\begin{align}\label{eq2}
  \bm{\tilde{y}} &=U(\bm{y}) \\
  \bm{\tilde{x}} & =g_s(\bm{\tilde{y}};\bm{\theta}) \notag
\end{align}
where $\bm{\tilde{y}}$ and $\bm{\tilde{x}}$ represent the latent variables with uniform noise added and its decoding reconstruction. Notation $U$ denotes adding uniform noise in training stage.
The difference between $\bm{\tilde{x}}$ and $\bm{x}$ is represented as distortion, and the entropy of $\bm{\tilde{y}}$ approximates the real code length.

To reduce the spatial redundancy in the latent variables $\bm{y}$, the work \cite{balle2018variational} proposed an auxiliary hyperprior network encoding its structural information $\bm{z}$. Formulated by
\begin{align}\label{eq3}
  \bm{z} & =h_a(\bm{y};\bm{\phi}_h) \notag \\
  \bm{\hat{z}} &=Q(\bm{z}) \\
  p_{\bm{\hat{y}|\hat{z}}}(\bm{\hat{y}|\hat{z}}) & \gets h_s(\bm{\hat{z}};\bm{\theta}_h) \notag
\end{align}
where $h_a$ and $h_s$ denote the encoder and decoder of this hyperprior network, and $\bm{\phi}_h$ and $\bm{\theta}_h$ correspond to their trainable parameters.

\begin{figure}
  \centering
  \includegraphics[width=9cm]{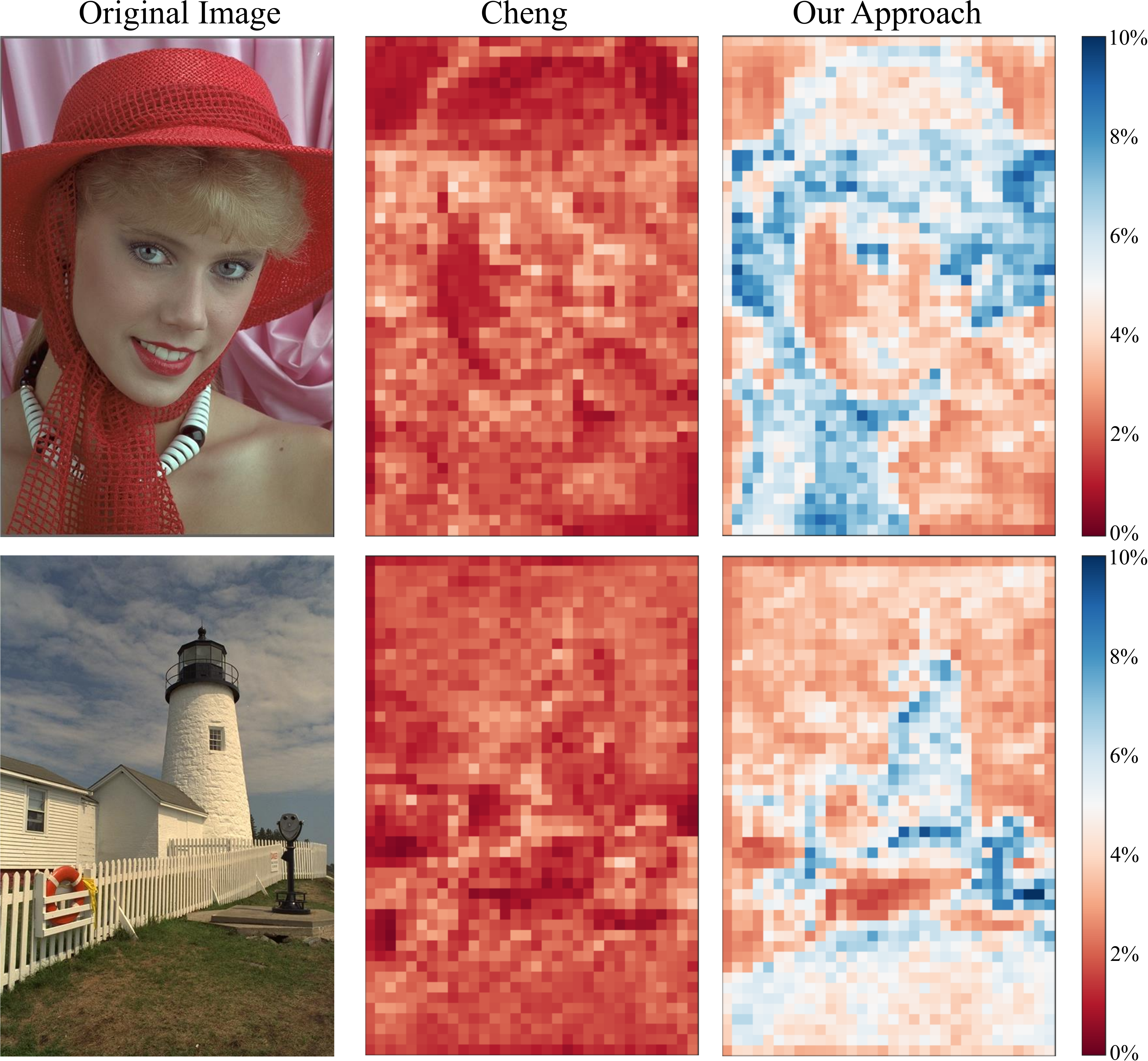}
  \caption{Average of the minimum weights of the GMM along the channel dimension (model using the MS-SSIM loss with $\lambda=12$). This average is very small in Cheng's method, revealing that the ternary GMM approximately degenerates into a binary model. In contrast, the minimum weight Gaussian model of our method still has a larger proportion in the region of complex textures, thus showing that our method makes better use of the GMM's ability to model the data.}\label{visual}
\end{figure}

\begin{figure*}
  \centering
  \includegraphics[width=18cm]{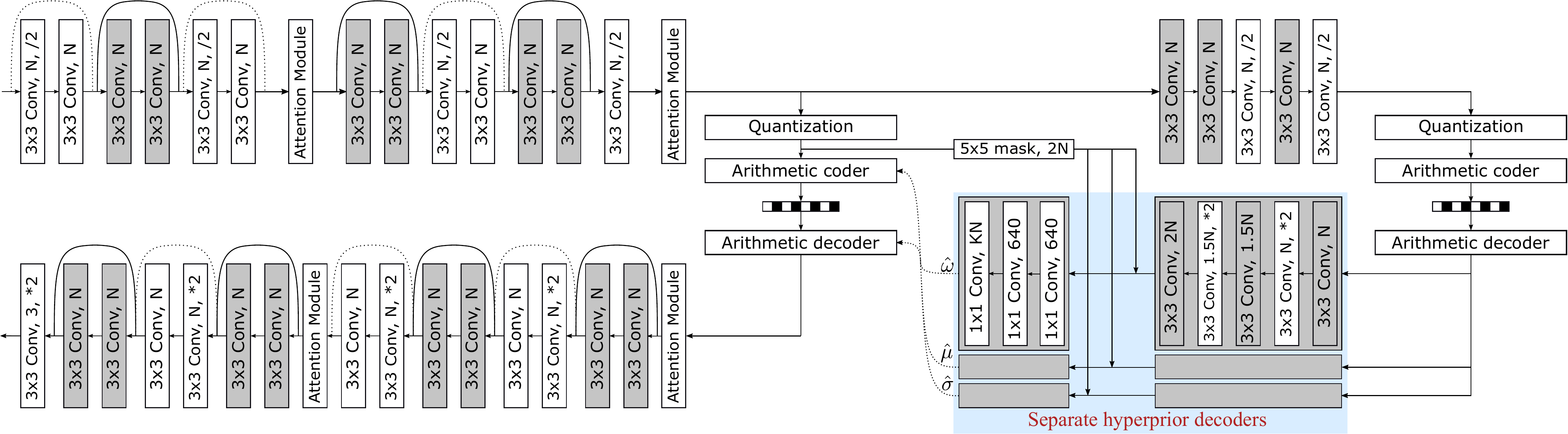}
  \caption{Network architecture.}\label{framework}
\end{figure*}

\subsection{Separate Hyperprior Decoders}
To enhance modeling capabilities for the prior $p_{\hat{\bm{y}}|\hat{\bm{z}}}(\hat{\bm{y}}|\hat{\bm{z}})$,
The work of Cheng \textit{et al.} \cite{cheng2020learned} proposed to use GMM, which contains three parameters of different physical significance, weight $\hat{\bm{\omega}}$, mean $\hat{\bm{\mu}}$ and variance $\hat{\bm{\sigma}}$.
\begin{equation}\label{eq5}
  p_{\hat{\bm{y}}|\hat{\bm{z}}}(\hat{\bm{y}}|\hat{\bm{z}}) \sim  \sum_{k=1}^{K}\hat{\bm{\omega}}^{(k)}N(\hat{\bm{\mu}}^{(k)}, \hat{\bm{\sigma}}^{2(k)})
\end{equation}
These parameters are obtained from the entropy parameter network $f$. And the hyper parameter $K$ denotes the number of Gaussian models in the GMM, which is set to 3 in both our and Cheng's model.
\begin{equation}\label{eq6}
  \hat{\bm{\omega}},\hat{\bm{\mu}},\hat{\bm{\sigma}}
   = f(c_m(\left\langle\hat{\bm{y}}\right\rangle),h_s(\hat{\bm{z}};\bm{\theta}_h))
\end{equation}
Function $c_m$ denotes the context model and the $\left\langle\hat{\bm{y}}\right\rangle$ denotes the already decoded subset of $\hat{\bm{y}}$ \cite{lee2018context}.
The 2-nd column in Fig. \ref{visual} demonstrates the impact of employing this strategy on the data modeling capabilities of the model.
Note that weight $\hat{\bm{\omega}}$ has a total of five dimensions, which are batch size, height, width, channel, and $K$ in order. We first take the minimum value of $\hat{\bm{\omega}}$ in the last dimension (i.e., $K$), and then take the average of the minimum value in the dimension of channel.
The value of this average can express the modeling ability of the GMM output from the hyperprior decoder. For example, this value of 0 is equivalent to the degradation of the GMM from a ternary model to a binary model. A similar situation occurs in Cheng, where a large number of averages are within 2\%. This means that the other two components occupy 98\% of the weight of GMM and the GMM degrades to some extent, thus leading to the inability of the model to model the data. This is probably caused by the decoding network's compromise among the three parameters. To avoid this entanglement of different parameters from a single network output, we use three separate hyperprior decoders and entropy parameter networks to decode the parameters here. In fact, the increase in complexity is limited because the tensor processed by the hyper model is downsampled several times.
\begin{align}\label{eq7}
  \hat{\bm{\omega}}
   = f_{\hat{\bm{\omega}}}(c_m(\left\langle\hat{\bm{y}}\right\rangle), h_{s(\hat{\bm{\omega}})}(\hat{\bm{z}};\bm{\theta}_{h(\hat{\bm{\omega}})})) \notag \\
  \hat{\bm{\mu}}
   = f_{\hat{\bm{\mu}}}(c_m(\left\langle\hat{\bm{y}}\right\rangle),
   h_{s(\hat{\bm{\mu}})}(\hat{\bm{z}};\bm{\theta}_{h(\hat{\bm{\mu}})})) \\
  \hat{\bm{\sigma}}
   = f_{\hat{\bm{\sigma}}}(c_m(\left\langle\hat{\bm{y}}\right\rangle),
   h_{s(\hat{\bm{\sigma}})}(\hat{\bm{z}};\bm{\theta}_{h(\hat{\bm{\sigma}})})) \notag
\end{align}
The 3-rd column in Fig. \ref{visual} of our case shows that the degradation phenomenon has been well improved.
The value of this average is more evenly distributed between 0 and 10\%.
In regions with relatively simple image textures, our model also degenerates into a binary Gaussian distribution, implying that the data itself may not need a complex distribution to be modeled. In contrast, in regions with complex image textures, such as the woman's hair and the lighthouse, our model uses a more complex ternary Gaussian distribution, which is more reasonable to model complex data distributions. This comparison visually demonstrates how our proposed method improves the performance of the original GMM approach in Cheng's work.

\subsection{Network Architecture and Training}
As shown in Fig. \ref{framework}, we use a network structure similar to Cheng\cite{cheng2020learned}, which employs the attention mechanism and cascaded residual blocks. The difference is that we propose to use separate hyperprior decoders in this framework. The decoded hyper latent code is fed to the three separate hyperprior decoders for decoding, and the obtained tensor is concatenated with the output of the context model and fed to the entropy parameter network to yield $\hat{\omega}$, $\hat{\mu}$ and $\hat{\sigma}$, respectively.

In training, the Lagrangian multiplier-based rate distortion loss of our model is
\begin{align}\label{eq8}
  L =& E_{\tilde{\bm{y}}, \tilde{\bm{z}} \sim q(\tilde{\bm{y}},\tilde{\bm{z}}|{\bm{x}})}
  [-log_2(p_{\tilde{\bm{y}}|\tilde{\bm{z}}}(\tilde{\bm{y}}|\tilde{\bm{z}}))
  -log_2(p_{\tilde{\bm{z}}|\bm{\psi}}({\tilde{\bm{z}}|\bm{\psi}}))] \notag \\
  & + \lambda \cdot D(\bm{x}, \tilde{\bm{x}})
\end{align}
where $q(\tilde{\bm{y}},\tilde{\bm{z}}|{\bm{x}})$ denotes the variational posterior in the autoencoder.  Model $p_{\tilde{\bm{z}}|\bm{\psi}}({\tilde{\bm{z}}|\bm{\psi}})$ denotes the non-parametric, fully factorized density model \cite{balle2018variational} used to encode $\bm{z}$, which can be formulated by
\begin{equation}\label{eq9}
  p_{\tilde{\bm{z}}|\bm{\psi}}({\tilde{\bm{z}}|\bm{\psi}}) = \prod_{i}\left(p_{z_i|\psi^{(i)}}({\psi^{(i)}})*U(-\frac{1}{2},\frac{1}{2})\right)(\tilde{z}_i)
\end{equation}

\begin{table}[tbp]
  \centering
  \caption{BD-rate Performance and Coding Complexity(Anchor: Cheng)}
  \setlength{\tabcolsep}{1.6mm}{
    \begin{tabular}{c|c|c|c|c|c}
    \hline
    Dataset & PSNR  & MS-SSIM & VMAF  & $\Delta$EncT  & $\Delta$DecT \bigstrut\\
    \hline
    Kodak & -1.13\% & -3.48\% & -1.06\% & 104.47\% & 103.96\% \bigstrut\\
    \hline
    CLIC  & -1.07\% & -2.42\% & -2.21\% & 103.42\% & 101.72\% \bigstrut\\
    \hline
    HEVC\_ClassB & -1.40\% & -2.65\% & -0.87\% & 102.72\% & 101.33\% \bigstrut\\
    \hline
    HEVC\_ClassC & -2.65\% & -4.18\% & -1.09\% & 104.37\% & 103.68\% \bigstrut\\
    \hline
    HEVC\_ClassD & -2.58\% & -3.80\% & -3.03\% & 107.63\% & 104.89\% \bigstrut\\
    \hline
    HEVC\_ClassE & -3.89\% & -3.64\% & -5.17\% & 103.56\% & 102.65\% \bigstrut\\
    \hline
    Average & \textbf{-2.12\%} & \textbf{-3.36\%} & \textbf{-2.24\%} & \textbf{104.36\%} & \textbf{103.04\%} \bigstrut\\
    \hline
    \end{tabular}%
    \label{Cheng_compare}%
  }
\end{table}%

\begin{table*}[tbp]
  \centering
  \caption{Comparison of Coding Efficiency and Absolute Coding Time of Different Codecs(Anchor: AVIF)}
    \begin{tabular}{c|c|r|r|r|r|r|r|r|r|r|r}
    \hline
    \multicolumn{2}{c|}{Dataset} & \multicolumn{1}{c|}{JPEG} & \multicolumn{1}{c|}{WEBP} & \multicolumn{1}{c|}{AVIF} & \multicolumn{1}{c|}{BPG} & \multicolumn{1}{c|}{x265} & \multicolumn{1}{c|}{VVenC} & \multicolumn{1}{c|}{HM} & \multicolumn{1}{c|}{VTM} & \multicolumn{1}{c|}{Cheng} & \multicolumn{1}{c}{Proposed} \bigstrut\\
    \hline
    \multirow{2}[4]{*}{Kodak} & PSNR  & 123.74\% & 76.66\% & 0.00\% & -10.58\% & 28.21\% & -7.16\% & -9.35\% & -28.23\% & -27.10\% &
    \textbf{-28.30\%} \bigstrut\\
\cline{2-12}          & MS-SSIM & 86.92\% & 63.79\% & 0.00\% & -10.20\% & 26.93\% & -22.55\% & -18.26\% & -28.08\% & -55.90\% &
    \textbf{-57.61\%} \bigstrut\\
    \hline
    \multirow{2}[4]{*}{CLIC} & PSNR  & 119.08\% & 99.32\% & 0.00\% & -6.25\% & 26.59\% & -2.78\% & -9.45\% & -23.74\% & -32.40\% &
    \textbf{-33.70\%} \bigstrut\\
\cline{2-12}          & MS-SSIM & 95.66\% & 62.47\% & 0.00\% & -16.12\% & 8.10\% & -38.46\% & -22.28\% & -40.39\% & -65.70\% &
    \textbf{-66.81\%} \bigstrut\\
    \hline
    \multirow{2}[4]{*}{HEVC} & PSNR  & 100.01\% & 72.65\% & 0.00\% & -12.30\% & 28.68\% & -14.89\% & -9.96\% & -22.38\% & -26.47\% &
    \textbf{-27.69\%} \bigstrut\\
\cline{2-12}          & MS-SSIM & 71.04\% & 51.01\% & 0.00\% & -14.42\% & 21.30\% & -25.36\% & -18.81\% & -42.48\% & -58.94\% &
    \textbf{-60.55\%} \bigstrut\\
    \hline
    \multicolumn{2}{c|}{Average} & 99.41\% & 70.98\% & 0.00\% & -11.65\% & 23.30\% & -18.53\% & -14.68\% & -30.88\% & -44.42\% & \textbf{-45.78\%} \bigstrut\\
    \hline
    \multicolumn{2}{c|}{EncT(s)} & \textbf{0.11} & \textbf{0.11}  & 79.43  & 1.08  & 1.15  & 18.49  & 46.11  & 310.53  & 77.23  & 79.97  \bigstrut\\
    \hline
    \multicolumn{2}{c|}{DecT(s)} & \textbf{0.0003} & 0.0021 & 0.26  & 0.61  & 0.39  & 0.85  & 0.69  & 0.48  & 75.84  & 77.42  \bigstrut\\
    \hline
    \end{tabular}%
  \label{main}%
\end{table*}%

\section{Experiment}
\subsection{Experimental Setting}
\textbf{Training.} We trained our model with CLIC training dataset \cite{clic} containing approximately 1600 images, using MSE with $\lambda$ in the set \{0.0016, 0.0032, 0.0075, 0.015, 0.03, 0.045\} and MS-SSIM with $\lambda$ in the set \{3, 12, 40, 120\} as quality metrics for optimization. We named the model optimized with the MSE metric as Model\_MSE, and the models trained with MS-SSIM metric as Model\_MS-SSIM. Hyper parameter $N$ is set as 128 for the lower-rate models and set as 192 for the higher-rate models, following the setting in the work of Cheng \textit{et al.}\cite{cheng2020learned}.
We use a randomly selected and cropped subset of the training set as the validation set containing 48 $256\times256$ patches. The batch size was set to $8$ and 1.08M iterations were conducted for each model to reach stable results. The models were optimized using Adam \cite{kingma2015adam}. The learning rate was maintained at a fixed value of $1\times10^{-4}$ during training, and was reduced to $1\times10^{-5}$ for the last 80K iterations. We chose variance scaling initializer for the filter kernel and zeros initializer for the bias vector.
The CPUs and GPUs in all experiments are Intel Xeon Gold 6230 CPU @ 2.10GHz and Nvidia RTX 2080 Ti GPU, respectively.

\textbf{Evaluation.} We used Kodak dataset\cite{kodak}, CLIC Professional Validation dataset\cite{clic}, and HEVC test sequences \cite{ctc} to evaluate the robustness of our method. Note that the HEVC dataset contains some video sequences in YUV format. We used the multimedia processing tool FFmpeg to convert the 1-st frame of each sequence into a PNG image, and finally combined them into a new dataset. Bits per pixel (BPP) is used to measure the rate, while PSNR and MS-SSIM are used to measure the image quality. For the implementation of MS-SSIM, we choose to use the calculation method of TensorFlow \cite{tf_msssim}. The BD-rate \cite{bjontegaard2001calculation} is used to quantitatively compare the compression performance between different codecs. Compared with the Rate-Distortion curves, the advantage of BD-rate is that it can quantitatively show Rate-Distortion performance regardless of whether the bit rate difference between models is obvious or subtle. We use an excel template proposed in \cite{wang2011on} for BD-rate calculation based on piece-wise cubic interpolation. For a fair comparison with the traditional codecs, the encoding and decoding times of the learned image codec are tested under CPU-only conditions.

\subsection{Performance Evaluation}
\textbf{Rate-distortion Performance.}
As shown in the Table \ref{main}, we compare the proposed method with traditional codecs, including JPEG \cite{jpeg}, WEBP\cite{webp}, AVIF\cite{avif}, BPG\cite{bpg}, HEVC (HM-16.16, x265-3.0)\cite{hevc}, VVC (VTM-11.2, VVenC-1.1.0)\cite{vvc} and learned codec\cite{cheng2020learned}. Because for codec HEVC and VVC the input and output are ususlly in YUV format, so we use PIL library\cite{pillow} to realize the exchange of the RGB format and the YUV format of the images. The format is YUV420 for VVenC, because currently only this format is supported, while for HM, x265 and VVC  the format is YUV444. All images are encoded with coding structure of all intra (AI) for codec HEVC and VVC. We set a series of QP values for every traditional codec, and the value of each QP remains constant during the compression process. Finally we select reasonable results with the quality that is closest to the quality measured by our models, which makes the calculation of BD-rates robust. Compared to these codecs, our models optimized by PSNR and MS-SSIM both achieve the best performance under three different test datasets. To compare the differences with the learned codecs more intuitively, we use Cheng \cite{cheng2020learned} as the anchor and calculated the BD-rate reduction and the relative coding times as shown in the Table \ref{Cheng_compare}. In order to make performance evaluation more convincing, Table \ref{Cheng_compare} also include the test results under VMAF metric, which are the average of Model\_MSE and Model\_MS-SSIM for each dataset.
It can be seen that our method achieves BD-rate reduction of 2.12\%, 3.36\% and 2.24\% under PSNR, MS-SSIM and VMAF metrics, respectively. For the HEVC\_ClassE dataset, our model optimized by MSE achieves a BD-rate reduction of 3.89\%. For the HEVC\_ClassC and HEVC\_ClassD datasets, our model optimized by MS-SSIM achieves a BD-rate reduction of about 4\%. Our models achieve the highest BD-rate reduction of 5.17\% under VMAF metric for the HEVC\_ClassE dataset. Overall, our method achieves better RD performance compared to these traditional codecs and Cheng's method.

\begin{figure}[tbp]
\centering
    \subfigure[N=128]{
    \begin{minipage}[t]{0.5\linewidth}
    \centering
    \includegraphics[scale=0.235]{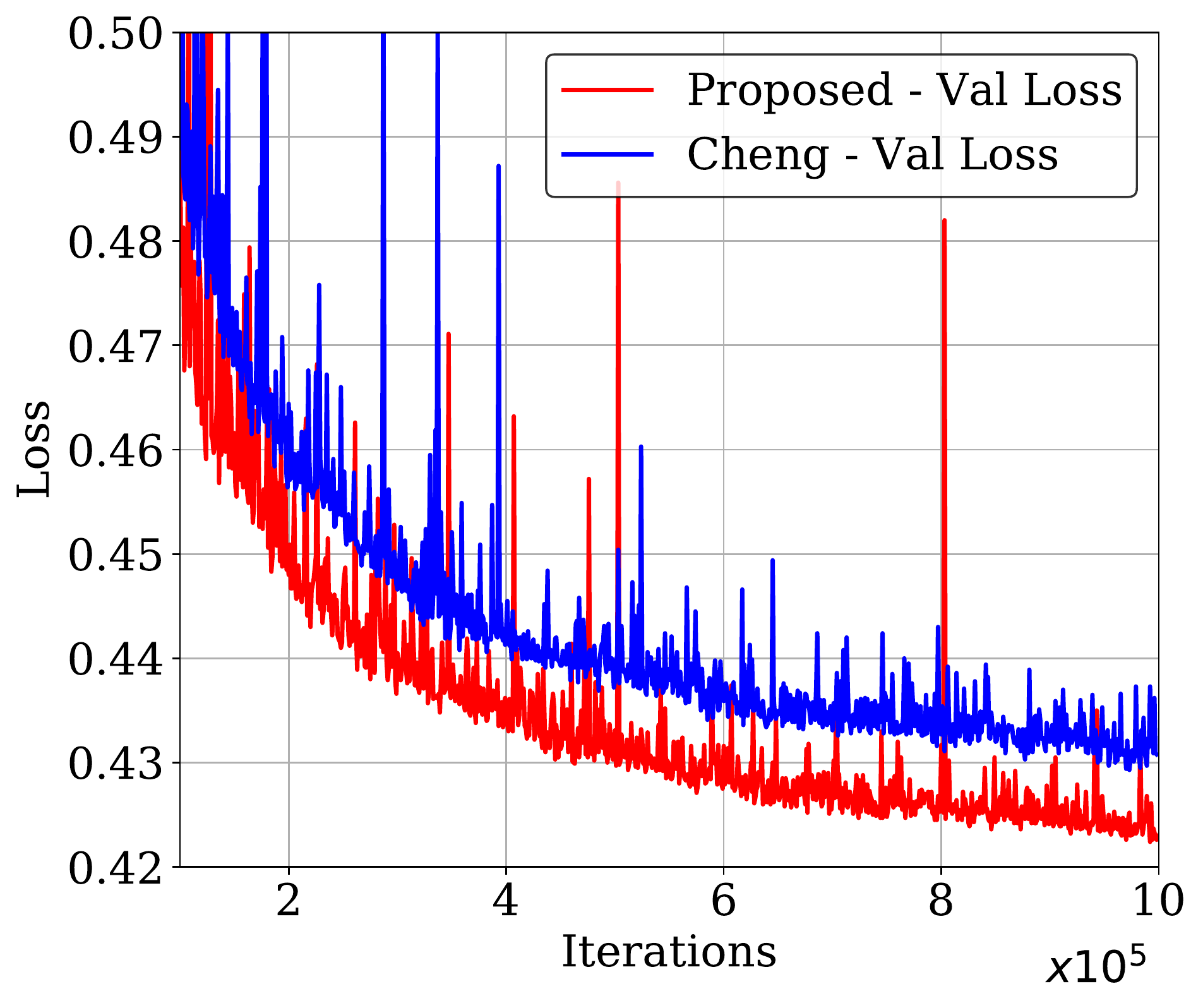}
    \end{minipage}%
    }%
    \subfigure[N=192]{
    \begin{minipage}[t]{0.5\linewidth}
    \centering
    \includegraphics[scale=0.235]{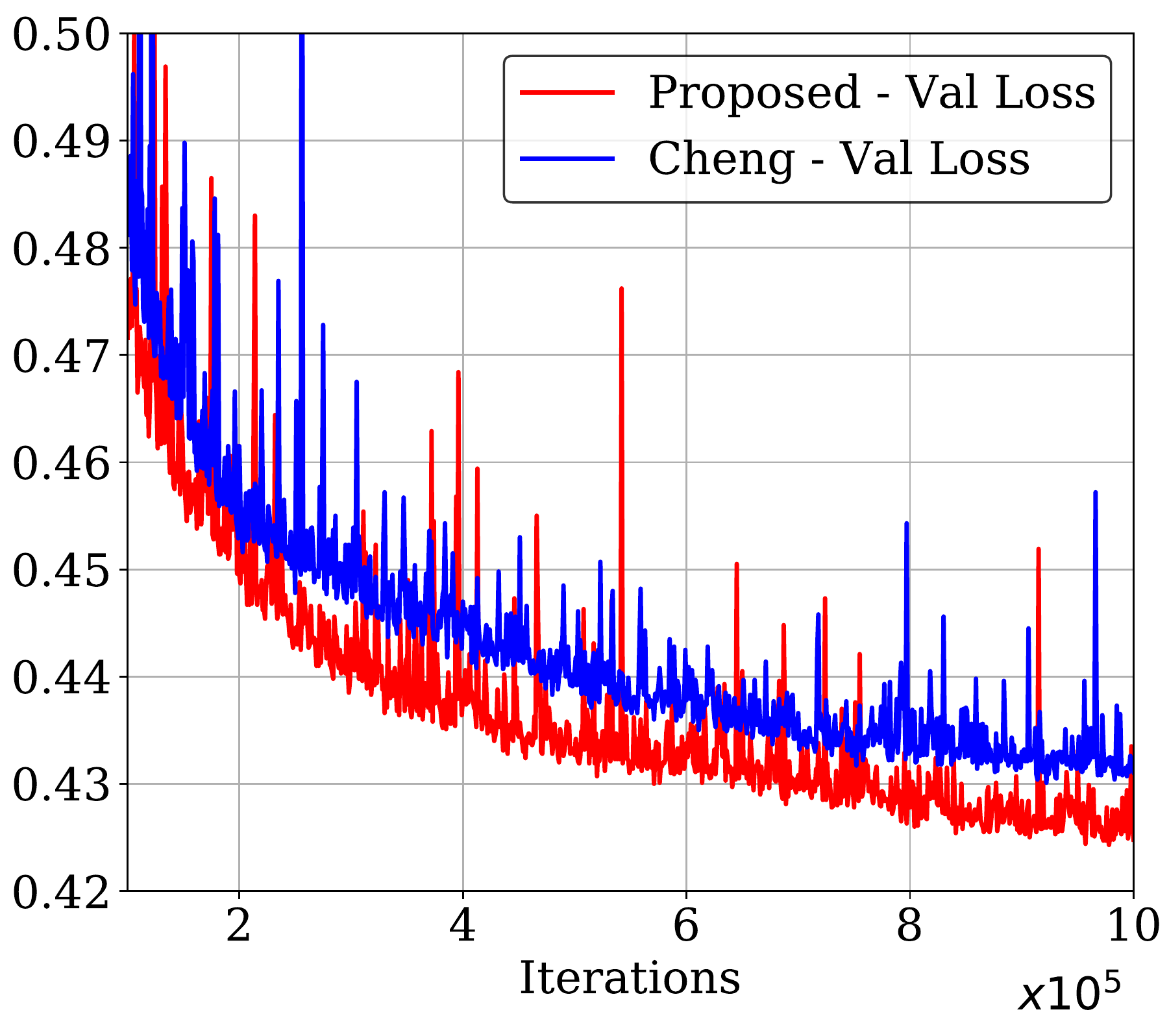}
    \end{minipage}%
    }%
\caption{Validation loss curves for models optimized with MS-SSIM($\lambda$=12).}\label{loss}
\end{figure}

\begin{figure}
  \centering
  \includegraphics[width=6cm]{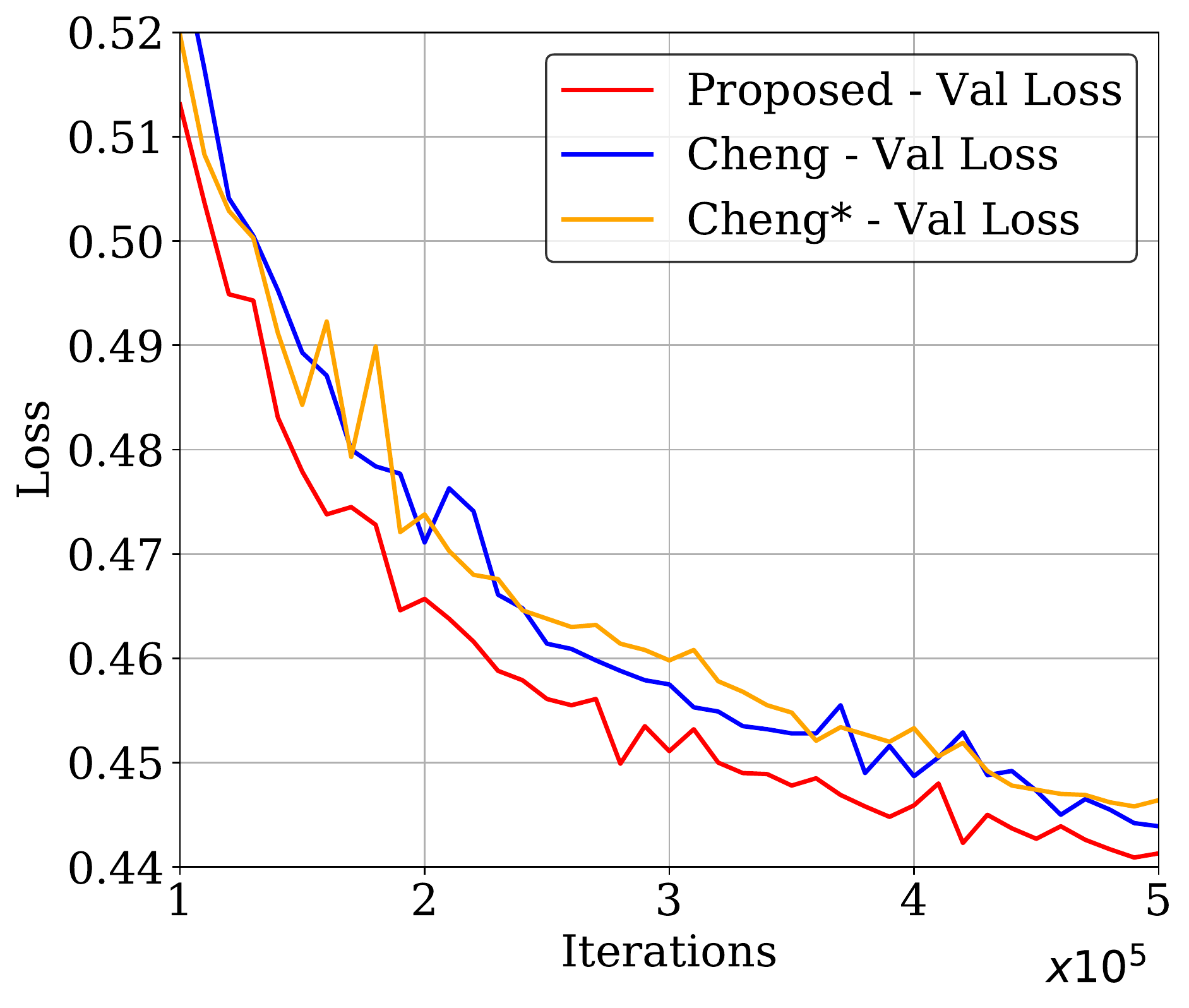}
  \caption{Validation loss curves for models optimized with PSNR($\lambda$=0.0032).}\label{complex}
\end{figure}

\textbf{Ablation Study.}
We strictly used the same dataset and training steps to train the proposed model and Cheng's model for a fair comparison, and their loss curves are shown in Fig. \ref{loss}. It shows that the proposed method has a smaller validation loss than the Cheng's method for models with N=128 and N=192. For the model with N=128, the proposed method improves the performance a little more than that of N=192. The rate-distortion results of the final converged model have been demonstrated in the previous subsection.

We also conducted an ablation study on the model complexity to demonstrate the effectiveness of our approach. We reserved the network structure of a single hyper decoder in Cheng's model, and purely increased the number of channels in each convolution layer in hyperprior decoder and entropy parameters, while the parameters in other modules were consistent. We name this network structure with higher complexity as Cheng*. The details about the individual layers with difference are shown in Table \ref{complex_ablation}. The total model size of Cheng* is slightly larger than the model we proposed. The loss curves of the three are shown in Fig. \ref{complex}. It can be observed that Cheng* cannot achieve a smaller loss than Cheng's model like the method we proposed. So in this case, the performance bottleneck doesn't lie in the amount of parameters, but the use of a single hyper decoder, which further proves the effectiveness of our proposed approach.

\begin{table}[tbp]
    \centering
    \caption{details about the individual layers with differences}
    \setlength{\tabcolsep}{1.8mm}{
    \begin{tabular}{c|c|c|c|c|c|c}
    \hline
    \multirow{2}{*}{Module}                                                       & \multirow{2}{*}{Layers} & \multirow{2}{*}{Kernel} & \multirow{2}{*}{Stride} & \multicolumn{3}{c}{Number   of Channels} \\ \cline{5-7}
                                                                                  &                         &                         &                         & Cheng       & Cheng*      & Proposed      \\ \hline
    \multirow{5}{*}{\begin{tabular}[c]{@{}c@{}}Hyperprior\\ Decoder\end{tabular}} & Conv1                   & 3x3                     & 1                       & 128         & 192         & 128   *3      \\ \cline{2-7}
                                                                                  & Conv2                   & 3x3                     & 2                       & 128         & 256         & 128   *3      \\ \cline{2-7}
                                                                                  & Conv3                   & 3x3                     & 1                       & 192         & 256         & 192   *3      \\ \cline{2-7}
                                                                                  & Conv4                   & 3x3                     & 2                       & 192         & 384         & 192   *3      \\ \cline{2-7}
                                                                                  & Conv5                   & 3x3                     & 1                       & 256         & 512         & 256   *3      \\ \hline
    \multirow{3}{*}{\begin{tabular}[c]{@{}c@{}}Entropy\\ Parameters\end{tabular}} & Conv1                   & 1x1                     & 1                       & 640         & 1024        & 640   *3      \\ \cline{2-7}
                                                                                  & Conv2                   & 1x1                     & 1                       & 640         & 1024        & 640   *3      \\ \cline{2-7}
                                                                                  & Conv3                   & 1x1                     & 1                       & 1152        & 1152        & 384   *3      \\ \hline
    \multicolumn{4}{c|}{Total Size (MB)}                                                                                                                       & 142.3       & 192.2       & 191.5         \\ \hline
    \end{tabular}
    \label{complex_ablation}
    }
\end{table}

\begin{table}[tbp]
  \centering
  \caption{Comparison of GFLOPs of Cheng and the Proposed Model}
    \begin{tabular}{c|c|c|c|c}
    \hline
    \multirow{2}[4]{*}{Method} & \multicolumn{2}{c|}{Size: 768x512} & \multicolumn{2}{c}{Size: 1920x1080} \bigstrut\\
\cline{2-5}          & N=128 & N=192 & N=128 & N=192 \bigstrut\\
    \hline
    Cheng & 339.78  & 757.57  & 1798.02  & 4008.75  \bigstrut\\
    \hline
    Proposed & 348.10  & 771.62  & 1841.77  & 4082.42  \bigstrut\\
    \hline
    Ratio & 102.45\% & 101.85\% & 102.43\% & 101.84\% \bigstrut\\
    \hline
    \end{tabular}%
  \label{gflops}%
\end{table}%

\textbf{Complexity.}
The average absolute coding times of all datasets by different codecs are given in Table \ref{main}, and the relative coding complexities compared to Cheng are given in Table \ref{Cheng_compare}.
Compared with the best-performing traditional method VTM, the absolute encoding time of our model is only about 1/4 of that of VTM. We know that VTM is the reference software of VVC, and it does not accurately reflect the complexity of a real implementation. Therefore, we tested the performance and complexity of VVenC, which is a practical implementation for VVC. Although the encoding and decoding time of VVenC is shorter, the average BD-rate reduction of VVenC with AVIF as the anchor is only 18.53\%, while that of our proposed approach is 45.78\%. The performance difference between the two codecs is very significant.
Compared with learned codec \cite{cheng2020learned}, the average encoding and decoding complexity of our model increases by 4.36\% and 3.04\%, respectively.
To quantitatively compare the complexity, we computed the FLOPs of these models as shown in Table \ref{gflops}. Compared to Cheng, the FLOPs of our model only increase by about 2\%. Note that the tensor processed by the hyper coding loop undergoes 4 downsampling operations in the main coding loop, indicating that the height and width of the processed tensor are 1/16 of the main loop, so the main model accounts for the main complexity of the overall model. Although the use of multiple hyper codecs has brought about a non-negligible increase in the amount of parameters and model size as shown in Table \ref{complex_ablation}, the impact on coding time and FLOPs is very limited.

\begin{figure}
  \centering
  \includegraphics[width=8.7cm]{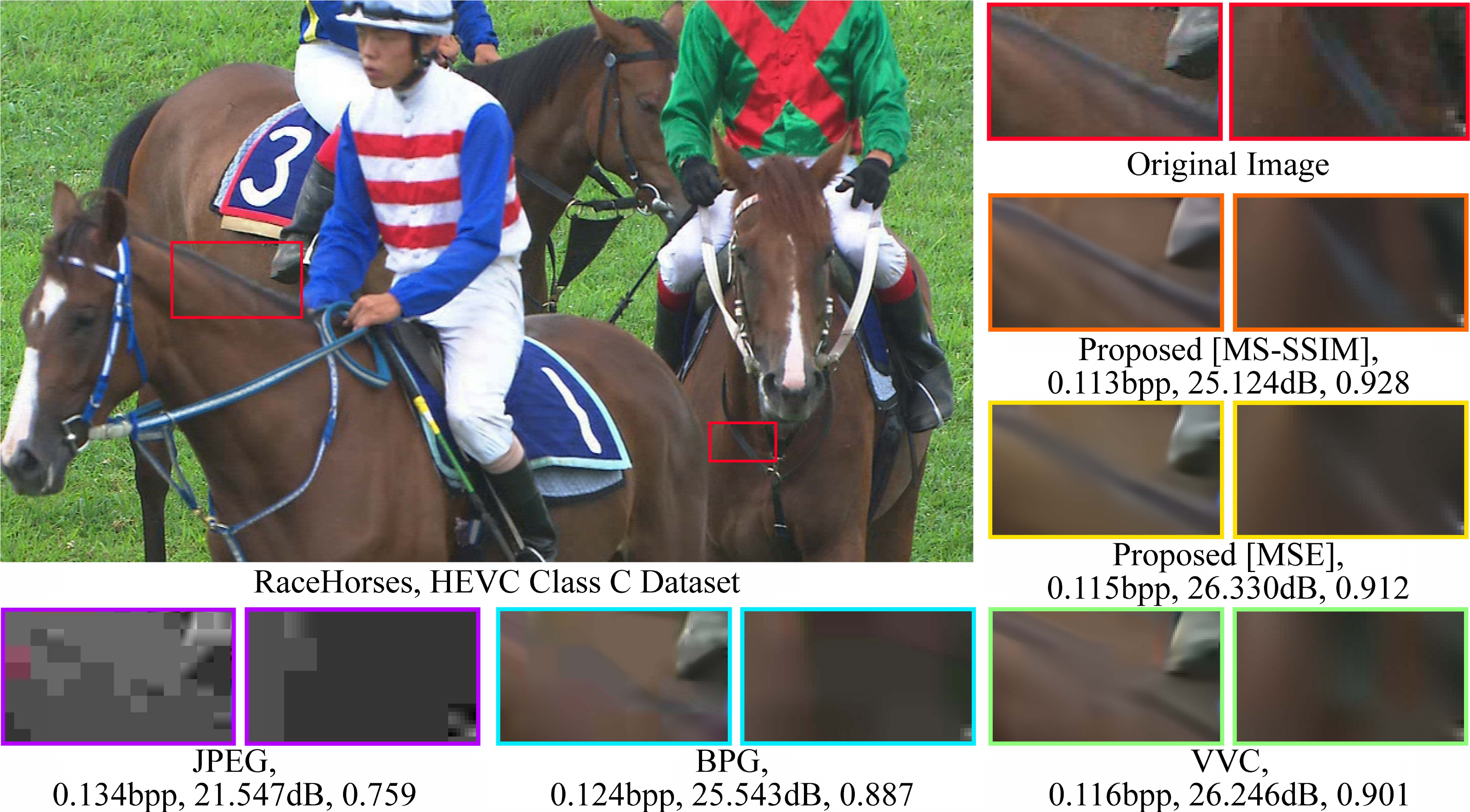}
  \caption{Visualization comparison of compressing the RaceHorses from HEVC test sequences.}\label{subjective}
\end{figure}

\textbf{Subjective Quality Evaluation.} We picked the RaceHorses in HEVC\_ClassC for a subjective quality comparison as shown in Fig. \ref{subjective}. At about 0.12bpp, the horse's mane and belts show a certain degree of distortion in VTM, and their quality is visibly worse in BPG and JPEG. In contrast, our model optimized by MSE retains the horse's mane. For our model optimized by MS-SSIM, both textures of the horse's mane and belts are well preserved and achieve a good subjective quality performance.

\section{Conclusion}
For variational autoencoder-based image compression, a direct and effective strategy to improve the model performance is proposed in this paper. By using separate hyperprior decoders for parameters of different physical significance in GMM, the value of the minimum weight in the GMM is improved. This results in improving the model's ability to compress complex images by generating more complex distributions to model it.
Compared to previous work of Cheng, our method achieves BD-rate gains of 2.12\%, 3.36\% and 2.24\% in terms of PSNR, MS-SSIM and VMAF metrics, respectively, while the cost of it to the coding time and FLOPs is negligible.

\bibliographystyle{IEEEtran}
\bibliography{reference}
\end{document}